\documentclass{Interspeech2024}

\usepackage[numbib]{tocbibind}
\usepackage[round]{natbib}
\interspeechcameraready

\title{Whispering in Norwegian: Navigating Orthographic and Dialectic Challenges}

\name[]{Per E}{Kummervold}
\name[]{Javier}{de la Rosa}
\name[]{Freddy}{Wetjen}
\name[]{Rolv-Arild}{Braaten}
\name[]{Per Erik}{Solberg}

\address{
  National Library of Norway  
  }
\email{per.kummervold@nb.no, versae@nb.no, freddy.wetjen@nb.no, rolv.braaten@nb.no, per.solberg@nb.no}

\keywords{speech recognition, language models, whisper}

\begin{document}

\maketitle

\begin{abstract}
    
This article introduces NB-Whisper, an adaptation of OpenAI's Whisper, specifically fine-tuned for Norwegian language Automatic Speech Recognition (ASR). We highlight its key contributions and summarise the results achieved in converting spoken Norwegian into written forms and translating other languages into Norwegian. We show that we are able to improve the Norwegian Bokmål transcription by OpenAI Whisper Large-v3 from a WER of 10.4 to 6.6 on the Fleurs Dataset and from 6.8 to 2.2 on the NST dataset. 

\end{abstract}

\section{Introduction}

Automatic Speech Recognition (ASR) holds the promise of changing the way we interact with technology by enabling machines to process human speech. 

Early Norwegian speech recognition research initially focused on limited vocabularies suitable for telephone applications, confronting challenges like compound words and varied pronunciations of numbers \citep{SvendsenPHH89,Paliwal1992OnTU}. Funded by the European Union at the start of the century, subsequent projects expanded the scope to more complex linguistic elements, contributing to valuable datasets and technical advancements using hidden Markov models and Mel Frequency Cepstral Coefficients \citep{amdal1995tabu,Hge1997EuropeanSD}. However, these systems, often utilising the Hidden Markov Model Toolkit \citep{young1994htk}, were limited in handling open-ended recognition and struggled with out-of-vocabulary words or real conversations. The introduction of newer datasets in recent years has led to systems with improved performance, setting the stage for the Wav2Vec work described in \cite{delarosa2023boosting} that seeks to address these longstanding challenges in Norwegian speech recognition.

OpenAI's Whisper \citep{radford2023robust} represents a significant deviation from the traditional ASR models. Unlike the standard approach of unsupervised pretraining followed by finetuning on a verbatim dataset mapping phonemes to graphemes, Whisper is pretrained on a vast, loosely aligned corpus of subtitles. This method has yielded impressive results, particularly in English, by not only handling capitalization and punctuation in a single step but also producing transcriptions that closely resemble written natural language. This approach effectively translates the spoken language domain into written text.

However, Whisper's effectiveness diminishes when dealing with Norwegian, a language characterised by its rich dialectical diversity and two written standards: Bokmål and Nynorsk. To address this limitation, we developed NB-Whisper, a specialised adaptation that provides tailored recognition for both Norwegian Bokmål and Nynorsk, as well as translations into English.

While our primary aim for the project is on enhancing ASR for Norwegian, the insights and methodologies developed through NB-Whisper have broader implications. Our work aims to contribute to the general body of knowledge in ASR, providing strategies and frameworks that can be adapted for other languages, especially those with similar linguistic complexities. By pushing the boundaries in Norwegian ASR, we seek to pave the way for innovations and improvements in speech recognition technologies globally.

\section{Language Variability and ASR Challenges}

Norwegian has two written standards with an equal official status: Bokmål and Nynorsk. Bokmål is close to the Oslo dialect and historically influenced by Danish. Nynorsk, written by about 13\% of Norwegian pupils in 2023 \citep{statisticsnorway2023}, is close to a number of rural dialects, particularly on the west coast. The two written standards differ mostly in inflectional forms and vocabulary, but there are also some syntactic differences. In addition, the orthographic norm of each written standard allows for a significant degree of variation compared to other European languages, mirroring the variation found in dialects. For example, all feminine nouns in Bokmål such as “lua”, ‘the hat’, can also be written in the masculine “luen”. In Nynorsk, infinitives can end on -e or -a, as in “å elske” or “å elska”, ‘to love’. 

The spoken dialects of Norwegian, while mutually intelligible, differ substantially in grammar, pronunciation and vocabulary. For example, the interrogative pronoun “who”, written “hvem” in Bokmål and “kven” in Nynorsk, is recorded with 38 distinct pronunciations in the Nordic Dialect Corpus \citep{johannessen-etal-2012-nordic}. There is no official norm of spoken Norwegian, and dialects are widely used, even in official settings such as in parliament and on the news.

ASR systems for Norwegian need to be able to handle the substantial dialect variation to be useful in real world applications. Also, they should be able to transcribe in both written standards. Written transcripts such as meeting notes, subtitles, or parliamentary interventions, are usually expected to render the speech consistently in Nynorsk or Bokmål, regardless of the dialect of the speaker, so the written standard should not depend on the speaker's dialect. Moreover, ASR test sets need to contain transcriptions in both written standards, from speakers of different dialects, and with metadata about dialects in order to give a realistic picture of ASR performance.

\section{Model Architecture and Training}

Whisper utilises an encoder-decoder Transformer architecture, with up to 30 second audio chunks re-sampled to 16,000 Hz and transformed into an 80-channel Mel spectrogram. For the latest Large-v3, this is expanded to a 128-channel Mel spectrogram \citep{openai2023whisper}. Feature normalisation and a two-layer convolutional encoder with GELU activation are applied, followed by sinusoidal position embeddings and transformer blocks. The decoder mirrors the encoder's structure, employing learned position embeddings. Byte-level BPE tokenization from GPT-2 is adapted for multilingual support \citep{radford2023robust}.

In our NB-Whisper project, we keep the core architecture of Whisper to be able to initiate training from the released checkpoint. Although the original PyTorch training code was not released by OpenAI, a collaborative effort with HuggingFace led to an alternative implementation in the Transformers library. This has also been adapted for Jax. The project participated in developing and open-sourcing training scripts for TPU-v4-pods, enabling dynamic changes to the training data during runtime \citep{nlnorway2024nostram}.

Our approach involves modifying some hyperparameters and training procedures. We increased the batch size to 1024 for all model sizes and scaled the learning rate accordingly. Warmup steps were slightly increased since we initialised from pretrained weights. Inspired by the Distil-Whisper work in \cite{gandhi2023distilwhisper} and their regularization practices, we also adopted a reduced setting for the weight decay with regards to the original OpenAI's Whisper implementation. We found that adding BPE dropout (increased to 0.2) was beneficial. Lastly, we did not implement the stochastic dropout used in OpenAI's Large v2 and v3. Instead we used activation dropout as a substitute, noting some but limited effect.

\begin{table}[h]
\centering
\begin{tabular}{lcc}
\textbf{Model} & \textbf{OpenAI Whisper} & \textbf{NB-Whisper} \\ \hline
Tiny           & $1.5 \times 10^{-3}$     & $6 \times 10^{-4}$  \\
Base           & $1 \times 10^{-3}$       & $4 \times 10^{-4}$  \\
Small          & $5 \times 10^{-4}$       & $2 \times 10^{-4}$  \\
Medium         & $2.5 \times 10^{-4}$     & $1 \times 10^{-4}$  \\
Large       & $2 \times 10^{-4}$       & $7 \times 10^{-5}$  \\
\end{tabular}
\caption{Learning rates.}
\label{tab:learning_rates}
\end{table}

\begin{table*}[!htbp]
\centering
\begin{tabular}{lccc}
\textbf{Hyperparameters} & \textbf{OpenAI Whisper} & \textbf{OpenAI Whisper Large v3} & \textbf{NB-Whisper} \\ \hline
Updates                  & 1,048,576                         & 655,360\footnotemark[1]                           & 200,000 + 50,000        \\
Batch Size               & 256                             & 1,024                             & 1,024                \\
Warmup Updates           & 2,048                            & 2,048                             & 10,000 / 5,000          \\
Max grad norm            & 1                               & 1                                & 1                   \\
Optimizer                & AdamW                           & AdamW                            & AdamW               \\
$\beta_1$                & 0.9                             & 0.9                              & 0.9                 \\
$\beta_2$                & 0.98                            & 0.98                             & 0.98                \\
$\epsilon$               & $10^{-6}$                       & $10^{-6}$                        & $10^{-6}$           \\
Weight Decay             & 0.1                             & 0.1                              & 0.01                \\
Weight Init              & Gaussian Fan-In                 & Gaussian Fan-In                  & OpenAI Whisper      \\
Learning Rate Schedule   & Linear Decay                    & Linear Decay                     & Linear Decay        \\
BPE Dropout              & 0                               & 0.1                              & 0.2                 \\
Stochastic Depth         & 0                               & 0                                & 0                   \\
Activation Dropout       & 0                               & 0                                & 0.1                 \\
\end{tabular}
\caption{Hyperparameters.}
\label{tab:hyperparameters}
\end{table*}

The training comprises two phases: an initial 200,000-step training, followed by dataset cleaning (see Section \ref{datasets}), and a final 50,000-step training. The latter phase is done on a cleaner dataset, and the main purpose of this training is to reduce hallucinations.

\section{Dataset}\label{datasets}

We train on the following sources:

\begin{enumerate}[label=(\alph*)] 
  \item \textbf{NRK Subtitles}. Subtitles from the Norwegian Broadcasting Corporation (NRK), known for their non-verbatim style, present a challenge for verbatim speech recognition due to significant deviations from spoken words. The subtitles are aligned using the timestamps from the subtitles. The data contains consistent notation for separating simultaneous speakers, continued sentences across timestamps, whether it is a recording of simultaneous texting, as well as some rarer notation for denoting the name of the speaker, the language spoken, credit to staff, etc. We clean this to get only the spoken text and combine it into longer segments. Due to the non-verbatim nature of the transcripts, a word-to-word transcription would typically have a word error rate (WER) around 30\% on this dataset. Additionally, we extract audio segments without speech from the same recordings. 

  \item \textbf{Audio Books}. Professional narrators read these audio books with minimal deviations from the text. For aligning the text we transcribed the text with NB-Wav2Vec2 1B\footnote{\url{https://huggingface.co/NbAiLab/nb-wav2vec2-1b-bokmaal}} by \cite{delarosa2023boosting}, and used the ideas from \cite{ljubesic-etal-2022-parlaspeech} for the forced alignment. Due to the existence of multiple correct spellings for words in Norwegian, even the best possible WER score hovers around 1-2\%. Still, using the words we found both in the NB-Wav2Vec2 transcription and the aligned text segments, we were able to extract chunks of text combined with corresponding sound clips.
  
  \item \textbf{The NST dataset}. This dataset was created by the now defunct Nordisk Språkteknologi. The dataset, which is distributed by the Language Bank at the National Library of Norway with an open licence, contains 540 hours of recordings of close to 1000 speakers with different dialectal backgrounds. As speakers tend to adapt their speech somewhat to the written language when they are reading out loud, the dataset has less evidence of dialectal grammar and vocabulary than datasets with spontaneous speech. When testing ASR systems on the NST dataset, we assume that the WER score will be approximately similar to audio books \citep{sprakbanken2023nst}.

  \item \textbf{The Stortinget Speech Corpus}. This is an open speech dataset created by the National Library of Norway in 2023 \citep{solberg-etal-2023-large}. It consists of around 5,200 hours of transcribed speech from Stortinget, the Norwegian parliament. The transcriptions are extracted from the official proceedings of Stortinget using a string matching algorithm adapted from \cite{ljubesic-etal-2022-parlaspeech}. 
\end{enumerate}

\begin{table*}[!htbp]
\centering
\begin{tabular}{lrr}
\textbf{Dataset}                  & \textbf{Stage 1 (hours)}         & \textbf{Stage 2 (hours)} \\ \hline
NRK - Subtitles                   & 16,518                           & 2,478                    \\  
NRK - No caption                  & 715                              & 312                      \\  
Audio Books                       & 2,461                            & 2,275                    \\  
The NST Dataset                   & 260                              & 490                      \\  
The Stortinget Speech Corpus      & 2,230                            & 523                      \\  
\textbf{Total}                    & \textbf{22,184}                  & \textbf{6,078}           \\  
\end{tabular}
\caption{Dataset size in hours. Augmentations (translations, timestamps etc) comes in addition.}
\label{tab:dataset_processing}
\end{table*}

To enhance our dataset, we augment the transcriptions with English translations. Most of the text originally in Norwegian Bokmål is translated into English using Google Translate, retaining any timestamp information. We observed that translation quality improves substantially with contextual information, such as translating entire programs and then splitting them using a designated token.

Our training process involves two stages. The first stage uses the complete dataset, and results in functioning models improving significantly on the OpenAI Whisper. We then use these models to run inference on the entire dataset. We identify various errors in the datasets, including misalignments and omissions. Instead of using a teacher-student approach where we train on the new output, we use the output to clean the dataset following several criteria:

\begin{itemize}
  \item \textbf{Fuzzy Matching for First and Last Words}. We ensure at least a partial match (80\% threshold) for the first and last words of each segment to filter out alignment errors.

  \item \textbf{Identifying Insertions}. We remove audio snippets where the target contains insertions not present in the audio, such as non-spoken transcriber comments. This is determined by finding target n-grams (longer than 3 words) that do not exist in any model predictions.

  \item \textbf{Detecting Omissions}. Snippets are deleted if they contain spoken phrases missing from the target text, identified by the absence of common n-grams (longer than 3 words) in all model predictions.

  \item \textbf{NER Analysis with BERT}. Since names often are inserted or spelled out especially in the parliament transcript, we perform Named Entity Recognition (NER) analysis using a BERT model. We use the number of named entities as a filter.
\end{itemize}

Any snippet that violates any of these filters, is deleted from the training set. We do not try to fix errors in the dataset.

\footnotetext[1]{Current numbers are for v2. It is reported that they have extended the dataset for v3 but OpenAI has to our knowledge not disclosed the number of trained steps.}

\section{Experimental Setup and Evaluation}

We evaluated our models using test sets from select datasets, specifically focusing on the Norwegian components:

\begin{itemize}
  \item \textbf{Fleurs Dataset}: This Google-curated, out-of-domain dataset comprises Norwegian Bokmål text snippets \citep{conneau2023fleurs}. It provides a valuable context for assessing model performance with unfamiliar content.

  \item \textbf{NST Dataset}: As detailed in Section \ref{datasets}, we used the test portion of this dataset distributed by \cite{sprakbanken2023nst}. Care was taken to avoid speaker overlap between this test set and our training and validation data. We note that both the NB-Whisper-models and the 
\end{itemize}

For evaluating model accuracy, we used the JiWER package, processing all texts to lowercase and removing punctuation prior to WER calculation. Our reporting of non-normalized scores aims to provide a clear and direct comparison across different datasets.

\section{Results and Discussion}

\begin{table}[!htbp]
\centering
\begin{tabular}{lccc}
\textbf{Model} & \multicolumn{2}{c}{\textbf{Bokmål}} & \textbf{Nynorsk} \\
               & Fleurs & NST & Common Voice \\
\hline
NB-Wav2Vec2 1B & 10.7 & 3.0 & 26.5 \\
OpenAI Large v1 & 13.2 & 12.7 & 51.7 \\
OpenAI Large v2 & 11.6 & 10.3 & 49.6 \\
OpenAI Large v3 & 10.4 & 6.8 & 30.0 \\
NB-Whisper Large (Ours) & 6.6 & 2.2 & 12.6 \\
\end{tabular}
\caption{Model performance comparison.}
\label{tab:model_performance}
\end{table}

\begin{table}[h]
\centering
\begin{tabular}{lcc}
\multicolumn{1}{c}{\textbf{Model}} & \textbf{OpenAI Whisper} & \textbf{NB-Whisper} \\ \hline
Tiny                               & 76.4                    & 15.2                \\
Base                               & 56.8                    & 11.5                \\
Small                              & 29.6                    & 8.3                 \\
Medium                             & 15.5                    & 7.2                 \\
Large                              & 10.4                    & 6.6                 \\
\end{tabular}
\caption{Norwegian Fleurs - Bokmål.}
\label{tab:norwegian_fleurs}
\end{table}

\begin{table}[h]
\centering
\begin{tabular}{lcc}
\textbf{Model} & \textbf{OpenAI Whisper} & \textbf{NB-Whisper} \\ \hline
Tiny           & 73.7                    & 8.1                 \\
Base           & 56.9                    & 4.9                 \\
Small          & 27.2                    & 3.1                 \\
Medium         & 14.6                    & 2.3                 \\
Large          & 6.8                     & 2.2                 \\
\end{tabular}
\caption{NST - Bokmål.}
\label{tab:nst}
\end{table}

\begin{table}[h]
\centering
\begin{tabular}{lcc}
\textbf{Model} & \textbf{OpenAI Whisper} & \textbf{NB-Whisper} \\ \hline
Tiny           & $>$100                    & 28.0                 \\
Base           & $>$100                   & 23.2                 \\
Small          & $>$100                     & 19.9                 \\
Medium         & 60.2                    & 17.0                 \\
Large          & 30.0                     & 12.6                 \\
\end{tabular}
\caption{Common Voice - Nynorsk.}
\label{tab:cv}
\end{table}

We are showing clear improvements from OpenAI Whisper for all the model sizes on a variety of datasets.

Table \ref{tab:model_performance} compares the performance of OpenAI and NB-Whisper models alongside NB-Wav2Vec2 models for both Bokmål and Nynorsk, highlighting their differences in architecture, size and approach. Whisper models, designed for robust multilingual transcription and translation, effectively handle noisy or varied speech data through extensive labeled data and context understanding techniques. In contrast, Wav2Vec2 models \cite{baevski202wav2vec} employ unsupervised pre-training from raw audio, followed by fine-tuning on labeled data. This makes them efficient for languages with scarce resources. However, the written output is based word by word on the input, and lack certain features typical for written language like capitalisation and punctuation. In comparison, the output from a Whisper model tends to more closely resemble written text in terms of quality.

Evaluating models quantitatively, such as by word-error-rate, presents certain challenges. Our focus is on scripted and read aloud text, where model comparability is highest. However, issues arise in normalization, like standardizing outputs amid speech diversity and lacking a universal dataset, complicating direct comparisons. Number normalization is notably difficult due to varying expressions and language-specific rules. In this study we did choose a light normalisation, mainly on punctuation and capitalization\footnote{A stronger normalization was applied in \cite{delarosa2023boosting} for the NB-Wav2Vec2 models (e.g., spelling out numbers), hence the difference in scores.}. Increasing the normalisation and standardisation, could have led to lower scores.


Tables \ref{tab:nst} and \ref{tab:cv} detail the performance of OpenAI Whisper and NB-Whisper models across various model sizes, comparing their effectiveness in recognizing Bokmål and Nynorsk languages. In the context of NST Bokmål dataset, the performance metrics indicate a clear trend: as the model size increases from Tiny (39M parameters) to Large (1550M parameters), the accuracy improves significantly for both OpenAI Whisper and NB-Whisper models, with NB-Whisper generally outperforming its OpenAI counterpart across all sizes. Specifically, the Tiny model shows a stark contrast in performance between the two, with NB-Whisper Tiny achieving a 8.1\% accuracy compared to OpenAI's Whisper's 73.7\%. This trend continues down to the Large model, where NB-Whisper achieves a 2.2\% word error rate, closely followed by OpenAI Whisper at 6.8\%.

In the Common Voice Nynorsk evaluation, the discrepancy in performance is even more pronounced, especially at smaller model sizes where OpenAI Whisper models report word error rates exceeding 100\%, indicating a significant challenge in handling Nynorsk. However, NB-Whisper achieves a remarkable 28.0\% WER score even at the Tiny model size. As the model size increases, performance improves dramatically, with the Large model achieving a 12.6\% error rate NB-Whisper and 30.0\% for OpenAI Whisper. This highlights the critical impact of model size on the accuracy of speech recognition tasks in different languages, particularly in less commonly supported languages like Nynorsk, where NB-Whisper models demonstrate a better ability to understand and transcribe the language accurately.

\section{Challenges and Limitations}

The evaluation of the model's handling of hallucinations and long text transcriptions was conducted qualitatively, primarily due to a shortage of Norwegian datasets specifically designed for such detailed quantitative assessments. For further developing Norwegian ASR it would be vital to develop such evaluation tools.

The model's architecture, while optimised for transcribing 30-second audio clips, presents a notable limitation in live transcription scenarios. The reliance on an autoregressive decoder is not ideal for doing real-time transcription. This suggests that alternative architectures might be more apt for applications requiring continuous, live transcription.

These challenges and limitations underscore the need for future advancements in model architecture and broader linguistic evaluations.

\section{Future Work}

Future improvements can be made in dataset cleaning. Our multi-stage approach, involving interim models, inherently risks discarding valuable data due to the fine line between high-quality and faulty data. Re-evaluating the entire dataset with improved tools developed during the project promises enhancements in data quality.

The high degree of orthographic variation in Norwegian, which permits multiple correct spellings, presents a substantial challenge. This variation can cause inconsistent model outputs due to the different acceptable spellings found in our corpus. Addressing this issue requires standardising the text within the corpus. A practical solution involves fine-tuning large language models (LLMs) to harmonise the corpus to a consistent style, significantly improving the quality of the dataset. This approach is particularly important for languages with a broad range of permissible orthographic variations, such as Norwegian.

\section{Conclusion}

The creation of NB-Whisper marks a step in advancing Automatic Speech Recognition (ASR) for Norwegian, adapting OpenAI's Whisper to meet the language's unique challenges. Our efforts in refining ASR capabilities have shown promising improvements in handling Norwegian's diverse dialects and written standards. However, we recognize that our project has limitations, especially in evaluating certain aspects like hallucinations and long text transcriptions. This is mainly because there are not enough detailed Norwegian datasets to thoroughly test these areas.

Looking ahead, the project underscores the importance of continuous research in ASR technology. While we have made strides in understanding and transcribing Norwegian speech, there is a clear pathway for further development, especially in model architecture and expanding linguistic adaptability. A particular focus should be on increasing the consistency in orthographic variation. This endeavour also opens possibilities for applying our learnings to other languages with similar complexities, contributing to the global progress of language technology.

\section{Funding and Acknowledgement}

Our research greatly benefited from the support provided by Google’s TPU Research Cloud (TRC), which generously supplied us with Cloud TPUs essential for our computational needs. We also extend our gratitude to Google Cloud for their support through Google credits, enabling us to carry out the necessary translations of our corpus into English. Special thanks are due to Sanchit Gandhi at Hugging Face for his substantial assistance in developing the TPU training scripts, a crucial component of our project. Thanks to Njaal Borch for providing the code enabling the initial alignment of the texts from NRK.

\bibliography{mybib}

\begin{thebibliography}{17}
\providecommand{\natexlab}[1]{#1}
\providecommand{\url}[1]{\texttt{#1}}
\expandafter\ifx\csname urlstyle\endcsname\relax
  \providecommand{\doi}[1]{doi: #1}\else
  \providecommand{\doi}{doi: \begingroup \urlstyle{rm}\Url}\fi

\bibitem[Amdal and Lj{\o}en(1995)]{amdal1995tabu}
Ingunn Amdal and Harald Lj{\o}en.
\newblock Tabu.0 - en norsk telefontaledatabase.
\newblock Technical Report~95, Scientific Report, 1995.

\bibitem[Baevski et~al.(2020)Baevski, Zhou, Mohamed, and Auli]{baevski202wav2vec}
Alexei Baevski, Henry Zhou, Abdelrahman Mohamed, and Michael Auli.
\newblock wav2vec 2.0: a framework for self-supervised learning of speech representations.
\newblock In \emph{Proceedings of the 34th International Conference on Neural Information Processing Systems}, NIPS'20, Red Hook, NY, USA, 2020. Curran Associates Inc.
\newblock ISBN 9781713829546.

\bibitem[Conneau et~al.(2023)Conneau, Ma, Khanuja, Zhang, Axelrod, Dalmia, Riesa, Rivera, and Bapna]{conneau2023fleurs}
Alexis Conneau, Min Ma, Simran Khanuja, Yu~Zhang, Vera Axelrod, Siddharth Dalmia, Jason Riesa, Clara Rivera, and Ankur Bapna.
\newblock Fleurs: Few-shot learning evaluation of universal representations of speech.
\newblock pages 798--805, 01 2023.
\newblock \doi{10.1109/SLT54892.2023.10023141}.

\bibitem[De~la Rosa et~al.(2023)De~la Rosa, Braaten, Kummervold, Wetjen, and Brygfjeld]{delarosa2023boosting}
Javier De~la Rosa, Rolv-Arild Braaten, Per~Egil Kummervold, Freddy Wetjen, and Svein~Arne Brygfjeld.
\newblock Boosting norwegian automatic speech recognition.
\newblock pages 555--564, May 2023.
\newblock URL \url{https://aclanthology.org/2023.nodalida-1.55}.

\bibitem[Gandhi et~al.(2023)Gandhi, von Platen, and Rush]{gandhi2023distilwhisper}
Sanchit Gandhi, Patrick von Platen, and Alexander~M. Rush.
\newblock Distil-whisper: Robust knowledge distillation via large-scale pseudo labelling, 2023.

\bibitem[H{\"o}ge et~al.(1997)H{\"o}ge, Tropf, Winski, van~den Heuvel, H{\"a}b-Umbach, and Choukri]{Hge1997EuropeanSD}
Harald H{\"o}ge, Herbert~S. Tropf, Richard Winski, Henk van~den Heuvel, Reinhold H{\"a}b-Umbach, and Khalid Choukri.
\newblock European speech databases for telephone applications.
\newblock \emph{1997 IEEE International Conference on Acoustics, Speech, and Signal Processing}, 3:\penalty0 1771--1774 vol.3, 1997.
\newblock URL \url{https://api.semanticscholar.org/CorpusID:7852982}.

\bibitem[Johannessen et~al.(2012)Johannessen, Priestley, Hagen, N{\o}klestad, and Lynum]{johannessen-etal-2012-nordic}
Janne~Bondi Johannessen, Joel Priestley, Kristin Hagen, Anders N{\o}klestad, and Andr{\'e} Lynum.
\newblock The {N}ordic dialect corpus.
\newblock In Nicoletta Calzolari, Khalid Choukri, Thierry Declerck, Mehmet~U{\u{g}}ur Do{\u{g}}an, Bente Maegaard, Joseph Mariani, Asuncion Moreno, Jan Odijk, and Stelios Piperidis, editors, \emph{Proceedings of the Eighth International Conference on Language Resources and Evaluation ({LREC}'12)}, pages 3387--3391, Istanbul, Turkey, May 2012. European Language Resources Association (ELRA).
\newblock URL \url{http://www.lrec-conf.org/proceedings/lrec2012/pdf/773_Paper.pdf}.

\bibitem[Ljube{\v{s}}i{\'c} et~al.(2022)Ljube{\v{s}}i{\'c}, Kor{\v{z}}inek, Rupnik, and Jazbec]{ljubesic-etal-2022-parlaspeech}
Nikola Ljube{\v{s}}i{\'c}, Danijel Kor{\v{z}}inek, Peter Rupnik, and Ivo-Pavao Jazbec.
\newblock {P}arla{S}peech-{HR} - a freely available {ASR} dataset for {C}roatian bootstrapped from the {P}arla{M}int corpus.
\newblock In Darja Fi{\v{s}}er, Maria Eskevich, Jakob Lenardi{\v{c}}, and Franciska de~Jong, editors, \emph{Proceedings of the Workshop ParlaCLARIN III within the 13th Language Resources and Evaluation Conference}, pages 111--116, Marseille, France, June 2022. European Language Resources Association.
\newblock URL \url{https://aclanthology.org/2022.parlaclarin-1.16}.

\bibitem[{OpenAI}(2023)]{openai2023whisper}
{OpenAI}.
\newblock Whisper large v3.
\newblock \url{https://huggingface.co/openai/whisper-large-v3}, 2023.
\newblock Accessed: Feb. 01, 2024.

\bibitem[Paliwal(1992)]{Paliwal1992OnTU}
Kuldip~K. Paliwal.
\newblock On the use of line spectral frequency parameters for speech recognition.
\newblock \emph{Digit. Signal Process.}, 2:\penalty0 80--87, 1992.
\newblock URL \url{https://api.semanticscholar.org/CorpusID:16553299}.

\bibitem[Radford et~al.(2023)Radford, Kim, Xu, Brockman, McLeavey, and Sutskever]{radford2023robust}
Alec Radford, Jong~Wook Kim, Tao Xu, Greg Brockman, Christine McLeavey, and Ilya Sutskever.
\newblock Robust speech recognition via large-scale weak supervision.
\newblock In \emph{Proceedings of the 40th International Conference on Machine Learning}, ICML'23. JMLR.org, 2023.

\bibitem[Solberg et~al.(2023)Solberg, Beauguitte, Kummervold, and Wetjen]{solberg-etal-2023-large}
Per~Erik Solberg, Pierre Beauguitte, Per~Egil Kummervold, and Freddy Wetjen.
\newblock A large {N}orwegian dataset for weak supervision {ASR}.
\newblock In Nikolai Ilinykh, Felix Morger, Dana Dann{\'e}lls, Simon Dobnik, Be{\'a}ta Megyesi, and Joakim Nivre, editors, \emph{Proceedings of the Second Workshop on Resources and Representations for Under-Resourced Languages and Domains (RESOURCEFUL-2023)}, pages 48--52, T{\'o}rshavn, the Faroe Islands, May 2023. Association for Computational Linguistics.
\newblock URL \url{https://aclanthology.org/2023.resourceful-1.7}.

\bibitem[{Statistics Norway}(2023)]{statisticsnorway2023}
{Statistics Norway}.
\newblock Table 03743: Pupils in primary and lower secondary school, by official form of norwegian 2023.
\newblock \url{https://www.ssb.no/en/statbank/table/03743}, 2023.
\newblock Accessed: Feb. 01, 2024.

\bibitem[Svendsen et~al.(1989)Svendsen, Paliwal, Harborg, and Husoy]{SvendsenPHH89}
Torbjørn Svendsen, Kuldip~K. Paliwal, Erik Harborg, and P.~O. Husoy.
\newblock An improved sub-word based speech recognizer.
\newblock In \emph{IEEE International Conference on Acoustics, Speech, and Signal Processing, ICASSP '89, Glasgow, Scotland, May 23-26, 1989}, pages 108--111. IEEE, 1989.
\newblock \doi{10.1109/ICASSP.1989.266375}.
\newblock URL \url{https://doi.org/10.1109/ICASSP.1989.266375}.

\bibitem[{The National Library of Norway}(2024)]{nlnorway2024nostram}
{The National Library of Norway}.
\newblock Nostram repository.
\newblock \url{https://www.github.com/NbAiLab/nostram}, 2024.
\newblock Accessed: Feb. 01, 2024.

\bibitem[{The Norwegian Language Bank}(2023)]{sprakbanken2023nst}
{The Norwegian Language Bank}.
\newblock Nst norwegian asr database.
\newblock \url{https://www.nb.no/sprakbanken/en/resource-catalogue/oai-nb-no-sbr-54/}, 2023.
\newblock Accessed: Feb. 01, 2024.

\bibitem[Young(1994)]{young1994htk}
Steve Young.
\newblock The htk hidden markov model toolkit: Design and philosophy.
\newblock \emph{Entropic Cambridge Research Laboratory, Ltd}, 2:\penalty0 2--44, 01 1994.

\end{thebibliography}

\end{document}